\PassOptionsToPackage{table}{xcolor}
\documentclass[sigconf]{acmart}
\copyrightyear{2026}
\acmYear{2026}
\setcopyright{cc}
\setcctype{by}

\acmConference[CIKM '26]
{Proceedings of the 35th ACM International Conference on Information and Knowledge Management}
{November 07--11, 2026}
{Rome, Italy}

\acmBooktitle{Proceedings of the 35th ACM International Conference on
Information and Knowledge Management (CIKM '26),
November 07--11, 2026, Rome, Italy}

\acmDOI{10.1145/3799682.3840968}
\acmISBN{979-8-4007-2539-5/2026/11}
\usepackage{amsthm}
\usepackage{amsmath}
\usepackage{algorithm}
\usepackage{enumitem}
\usepackage{picinpar}
\usepackage{lineno}
\usepackage{graphicx}
\usepackage[table]{xcolor}
\usepackage{algorithmicx}
\usepackage[noend]{algpseudocode}
\usepackage{subfigure}
\usepackage{multirow}
\usepackage{color}
\usepackage{enumitem}
\usepackage{hhline}
\usepackage[normalem]{ulem}
\usepackage{booktabs}
\usepackage{wrapfig}
\usepackage{cancel}
\usepackage{hyperref}
\usepackage{makecell}
\usepackage{balance}

\newtheorem{lemma}{Lemma}

\newtheorem{definition}{Definition}
\newtheorem{proposition}{Proposition}

\newcommand{\bL}{\ensuremath{\mathcal{L}}}

\newcommand{\bV}{\ensuremath{\mathcal{V}}}

\newcommand{\bN}{\ensuremath{\mathcal{N}}}
\newcommand{\bE}{\ensuremath{\mathcal{E}}}

\newcommand{\bM}{\ensuremath{\mathcal{M}}}

\newcommand{\bD}{\ensuremath{\mathcal{D}}}

\renewcommand{\vec}[1]{\ensuremath{\mathbf{#1}}}

\newcommand{\stitle}[1]{\vspace{1mm} \noindent {\bf #1}}

\newcommand{\method}[1]{\textsc{#1}}
\newcommand{\model}{\method{MolGA}{}}

\newcommand{\eat}[1]{}

\newcommand{\stkout}[1]{\ifmmode\text{\sout{\ensuremath{#1}}}\else\sout{#1}\fi}

\usepackage{amsmath,amsfonts,bm}

\def\eqref#1{equation~\ref{#1}}

\def\1{\bm{1}}

\DeclareMathAlphabet{\mathsfit}{\encodingdefault}{\sfdefault}{m}{sl}
\SetMathAlphabet{\mathsfit}{bold}{\encodingdefault}{\sfdefault}{bx}{n}

\usepackage{hyperref}
\usepackage{url}

\title{Molecular Knowledge Guided Prompt Adaptation for Pre-trained 2D Graph Encoders}

\author{Xingtong Yu}
\authornote{Both authors contributed equally to this research.}
\authornote{This work was completed while the author was a research scientist at Singapore Management University.}
\affiliation{%
  \institution{The Chinese University of Hong Kong}
  \city{Hong Kong}
  \country{China}
}
\email{xtyu@se.cuhk.edu.hk}

\author{Chang Zhou}
\authornotemark[1]
\affiliation{%
  \institution{University of Science and Technology of China}
  \city{Hefei}
  \country{China}
}
\email{zhouchang21sy@mail.ustc.edu.cn}

\author{Xinming Zhang}
\authornote{Corresponding authors.}
\affiliation{%
  \institution{University of Science and Technology of China}
  \city{Hefei}
  \country{China}
}
\email{xinming@ustc.edu.cn}

\author{Yuan Fang}
\authornotemark[3]
\affiliation{%
  \institution{Singapore Management University}
  \city{Singapore}
  \country{Singapore}
}
\email{yfang@smu.edu.sg}

\renewcommand{\shortauthors}{Xingtong Yu, Chang Zhou, Xinming Zhang, and Yuan Fang}

\begin{document}

\begin{abstract}
Pre-trained 2D graph encoders have achieved strong performance in molecular representation learning by capturing transferable topological patterns from large-scale unlabeled molecular graphs. However, downstream molecular applications often involve rich chemical and geometric information associated with atoms and bonds, which is not explicitly modeled by conventional 2D graph pre-training approaches. Existing molecular pre-training methods typically incorporate such knowledge during large-scale pre-training, requiring specialized architectures and substantial computational cost.
In this paper, we investigate how to enhance pre-trained 2D graph encoders through lightweight downstream prompt adaptation with molecular knowledge. We propose a molecular prompting framework that integrates heterogeneous molecular information into downstream graph learning without modifying the pre-trained encoder parameters. Specifically, molecular knowledge is first projected into the latent space of pre-trained topological representations to reduce the discrepancy between topology and molecular semantics. Based on the aligned representations, we further employ knowledge-conditioned prompt generators to produce adaptive prompts for different molecular components, enabling fine-grained and instance-aware downstream adaptation.
The proposed framework can be flexibly incorporated into existing 2D graph pre-training methods while maintaining strong parameter efficiency. Extensive experiments on molecular classification and molecular property prediction benchmarks demonstrate consistent improvements over representative graph pre-training and molecular representation learning baselines, particularly under low-resource settings.
\end{abstract}



\maketitle
\section{Introduction}\label{sec.intro}
Molecular graph representation learning has emerged as a mainstream technique in computational chemical and biomedical research \citep{gilmer2017neural,rong2020self}. Early studies primarily leveraged graph encoders, such as graph neural networks (GNNs) \citep{kipf2016semi,velivckovic2017graph,yu2023learning} and transformers \citep{ying2021transformers,yun2019graph}, to encode 2D topological structures of molecules, where nodes represent atoms and edges represent chemical bonds. However, their effectiveness relies heavily on large amounts of labeled data, which are expensive to obtain through labor-intensive wet-lab experiments or computationally demanding physics-based simulations. 
To overcome this limitation, graph pre-training methods have emerged as a promising solution. They typically follow a two-stage paradigm: first, the graph encoder learns task-agnostic and intrinsic properties from large-scale unlabeled graphs in a self-supervised manner; then, it is fine-tuned on downstream tasks using limited labeled data \citep{luong2023fragment,sypetkowski2024scalability,yang2024masking,mendez2024mole,yu2025gcot}.

\begin{figure*}[t]
\centering
\includegraphics[width=1\linewidth]{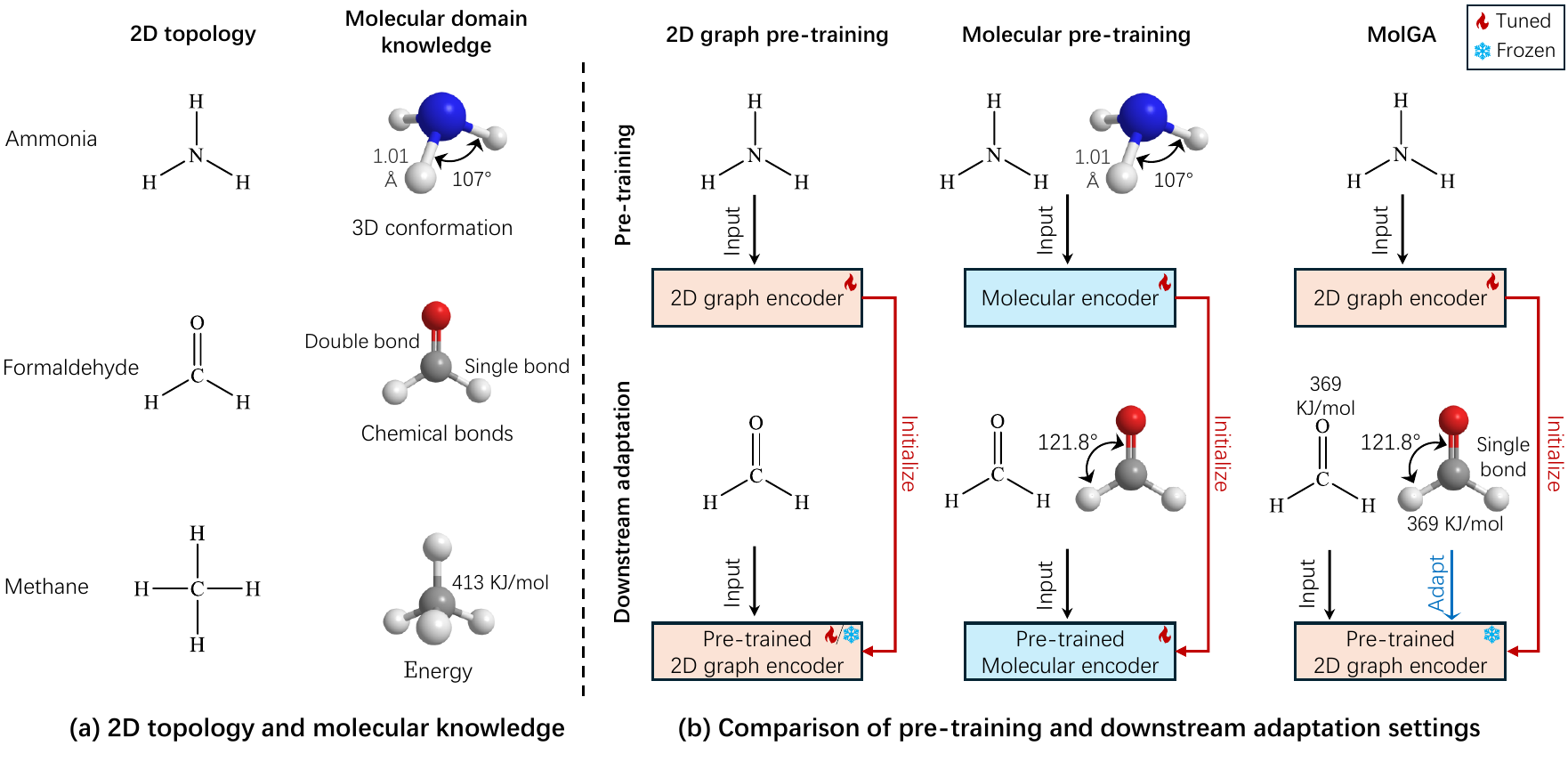}
\vspace{-8mm}
\caption{Motivation of \model. (a) 2D topological and molecular domain knowledge in molecules. (b) Comparison of pre-training and downstream adaptation settings in 2D pre-training methods, molecular pre-training approaches and \model.}
\label{fig.intro-motivation}
\end{figure*}

Beyond 2D topological structures, molecules contain a variety of molecular domain knowledge that characterizes the chemical and physical properties of submolecular instances (atoms and bonds), as illustrated in Fig.~\ref{fig.intro-motivation}(a). For example, 3D conformations \citep{schutt2017schnet,zaidi2022pre} capture the spatial arrangement of  atoms in three-dimensional space. Chemical bond types describe different bonding relationships between atoms, such as single and double bonds \citep{liu2021spherical,wang2022comenet}. Atom energy reflects the intrinsic chemical energy associated with each atom or bond \citep{zou2023deep,wang2025molspectra}.
Leveraging such rich information, recent studies have extended 2D graph learning methods by integrating molecular domain knowledge into graph encoders to enhance their expressiveness \citep{zaidi2022pre,wang2025molspectra}. Additionally, molecular domain knowledge has been incorporated into pre-training objectives \citep{liupre,fang2022geometry,yu2023multimodal}, leading to the development of molecular pre-training methods. Specifically, geometry-related knowledge (such as 3D conformations) should respect the physical $SE(3)$ symmetry:
molecular properties are invariant to global rotations and translations, and so should geometry-dependent downstream adaptations.

Current molecular pre-training approaches typically incorporate molecular domain knowledge during the pre-training phase \citep{liupre,fang2022geometry}, and the pre-trained model is further fine-tuned for downstream tasks involving the same type of domain knowledge.
However, a plethora  of well-established pre-trained 2D graph encoders already exist, demonstrating robustness and generalization with extensive empirical validation \citep{sypetkowski2024scalability,yang2024masking}. Leveraging existing pre-trained 2D graph encoders could significantly reduce costs compared with pre-training a new molecular encoder, which generally involves greater complexity and a larger number of parameters \citep{schutt2017schnet,liu2021spherical,wang2022comenet}. 
Moreover, existing molecular pre-training methods typically focus on a single type of domain knowledge, thereby overlooking the benefits of incorporating different forms of molecular information \citep{liumolecular, zhouuni, kim2023geotmi, wang2025molspectra}. These observations naturally raise a fundamental question: \textit{Given a pre-trained 2D graph encoder, can downstream adaptation flexibly integrate different types of molecular domain knowledge to enhance its performance in molecular applications?} 
In this work, we propose a \textbf{Mol}ecular \textbf{G}raph \textbf{A}daptation framework, \model, which adapts pre-trained 2D graph encoders to downstream molecular tasks by flexibly incorporating diverse molecular domain knowledge. A comparative overview of the pre-training and downstream adaptation settings across 2D graph pre-training, molecular pre-training, and \model\ is illustrated in Fig.~\ref{fig.intro-motivation}(b).
The realization of \model\ is non-trivial due to two key challenges.

First, \textit{how do we align pre-trained 2D topological representations with molecular domain knowledge for downstream adaptation?} Pre-trained 2D graph encoders typically capture flattened topological patterns, such as atom connectivity and graph motifs. In contrast, molecular domain knowledge differs across submolecular instances (atoms and bonds), with instance-specific chemical or physical characteristics that can vary among similar topological patterns. This discrepancy leads to conflict, rather than synergy, when the two sources of knowledge are na\"ively fused in downstream adaptation.
In \model, we employ a contrastive alignment strategy that bridges the gap between pre-trained topology and molecular domain knowledge. Specifically, for each atom and bond, we aim to maximize the similarity between its 2D topological representation and its domain-knowledge representation, while minimizing similarity across different instances.

Second, \textit{how can we leverage the aligned knowledge to adapt to fine-grained, instance-specific characteristics for molecular applications?} Existing molecular pre-training approaches typically leverage a one-size-fits-all adaptation (fine-tuning) strategy for downstream tasks \citep{liupre,fang2022geometry,zong2024deep,wang2025molspectra,yu2024survey}, disregarding the heterogeneity of submolecular instances. Within a molecule, different atoms and bonds exhibit heterogeneous topological, chemical and physical properties, reflecting their unique roles in the molecule. For example, in formaldehyde ($\text{CH}_2\text{O}$), the oxygen atom, due to its high electronegativity, renders the C=O bond highly polarized; the carbon atom bears a partial positive charge, making it a favorable site for nucleophilic attack; the C--H bonds are relatively stable, primarily serving to maintain the structural framework of the molecule \citep{xu2018analyte}.
Therefore, fine-tuning all submolecular instances uniformly overlooks such intrinsic variability, limiting the expressiveness and effectiveness of their downstream representations.
In \model, we introduce a parameter-efficient, instance-specific adaptation strategy. Inspired by conditional prompting \citep{zhou2022conditional,yu2024non,yu2024multigprompt,yu2024generalized}, \model\ employs lightweight conditional networks that condition on the aligned instance-level representations to generate a series of instance-specific tokens. The atom-specific tokens are used to adjust atom features before feeding them into the frozen pre-trained graph encoder, while bond-specific tokens are injected into the message passing process. By updating only the conditional networks while keeping the encoder frozen, \model\ enables fine-grained yet lightweight adaptation to downstream tasks.

In summary, the contributions of this work are fourfold. 
(1) We propose a contrastive alignment strategy to bridge the representational gap between pre-trained 2D topological knowledge and downstream molecular domain knowledge.
(2) Realizing that various submolecular instances exhibit distinct topological, chemical and physical characteristics, we introduce an instance-specific downstream adaptation mechanism using lightweight conditional networks. 
(3) We theoretically prove that \model\ is $SE(3)$-invariant, ensuring invariance to global rotations and translations when 3D conformations are involved.
(4) We conduct extensive experiments on a wide array of benchmarks, demonstrating that \model\ consistently outperforms state-of-the-art methods across various molecular applications.

\section{Related Work}
We briefly review related work on molecular graph representation and pre-training.

\stitle{Molecular graph representation learning.}
To model the 2D topological structure in molecules, GNNs \citep{kipf2016semi, velivckovic2017graph} and transformers \cite{yun2019graph,ying2021transformers} have become mainstream techniques. 
While GNNs aggregate information from local neighborhoods through message passing, Transformers leverage attention to capture interactions between nodes across the graph.
However, molecules are also associated with rich physical and chemical domain knowledge, such as 3D conformations and chemical bond types. Hence, recent studies have further incorporated molecular domain knowledge \citep{schutt2017schnet, liu2021spherical, wang2022comenet}, embedding such information directly into node representations.
Despite their success, both 2D graph and molecular encoders require substantial labeled data to train for each new task and dataset, limiting their generalizability and scalability.

\stitle{Molecular graph pre-training.}
To address the limitations of supervised methods, graph pre-training techniques have been developed to capture task-agnostic structural patterns in a self-supervised manner \citep{kipf2016variational, hu2020strategies, you2020graph, hu2020gpt, yu2024few}. These methods typically pre-train a 2D graph encoder to learn inherent properties from unlabeled graphs and then adapt (fine-tune) the encoder on downstream tasks \citep{luong2023fragment,sypetkowski2024scalability,yang2024masking,mendez2024mole}. 
While 2D graph pre-training approaches focus on topological structures, 
molecular-specific pre-training methods have been proposed to capture rich molecular domain knowledge. Some approaches focus on denoising or reconstructing the 3D geometry of molecules during pre-training \citep{liumolecular, zhouuni, kim2023geotmi, wang2025molspectra}, while others incorporate both 2D topological structures and molecular knowledge, including stereochemistry and atom-level interactions, into their pre-training objectives \citep{stark20223d, liupre, liu2023group, chen2023molecule, zong2024deep}.
However, existing molecular pre-training methods typically adopt a one-size-fits-all downstream adaptation for all instances, overlooking the distinct chemical and physical characteristics exhibited by different instances. Moreover, these methods require pre-training a molecular graph encoder with a significantly larger number of parameters compared to 2D graph encoders. 


\section{Preliminaries} \label{sec.preliminary}
In this section, we introduce related preliminaries and the scope of our work.

\stitle{Molecular graph.}
A 2D graph is defined as \( G = (\bV, \bE) \), where $\bV$ and $\bE$ denote the set of nodes (atoms) and edges (bonds), respectively. A molecular graph is defined as \( G = (\bV, \bE, \bM) \), which
further incorporates a set of molecular domain knowledge \( \bM = \{M_1, \ldots, M_K\}\). Each \( M_k \in \bM \) represents a type of domain knowledge on atoms or bonds. Moreover, each atom \( v_i \in \bV \) is associated with a feature vector \( \vec{x}_i \in \mathbb{R}^{d} \), and the entire atom feature matrix is denoted as \( \mathbf{X} \in \mathbb{R}^{|\bV| \times d} \). We write a collection of molecular graphs as \( \mathcal{G} = \{ G_1, G_2, \dots, G_N \} \).

\stitle{2D Graph encoder.}
 A typical implementation utilizes GNNs, which update each node's embedding by iteratively aggregating information from its local neighborhood. 
Let \( \vec{H}_\bV^l \in \mathbb{R}^{|\bV| \times d} \) be the node embedding matrix at the \( l \)-th layer, where each row \( \vec{h}_i^l \) represents the embedding of node \( v_i \). In layer \( l \), the forward propagation is given by
 $   \vec{h}^l_v = \mathtt{MP}(\vec{h}^{l-1}_v, \{\vec{h}^{l-1}_u : u\in\bN_v\}; \theta^l)$,
where \( \mathcal{N}_v \) is the neighboring atoms of \( v \), \( \mathtt{MP}(\cdot) \) is the message-passing function and \( \theta^l \) denotes the learnable parameters of the \( l \)-th layer. In the first layer, the input feature vector $\vec{x}_v$ serves as the initial embedding \( \vec{h}^0_v \), and the final output of the GNN after \( L \) layers is denoted  as \( \vec{h}_v \), which is a row of the node embedding matrix \( \vec{H}_\bV \). 
We denote the overall encoding process over \( L \) layers as
\begin{align}\label{eq.gnn}
    \vec{H}_\bV = \mathtt{GE}(\vec{X}, G; \Theta),
\end{align}
where \( \mathtt{GE}(\cdot) \) denotes the 2D graph encoder, and \( \Theta = \{\theta^1, \theta^2, \dots, \theta^L\} \) represents the set of trainable parameters across all layers. 
For the edge embedding matrix \( \vec{H}_\bE \),  each row \( \vec{h}_{(u,v)\in \bE} \) is computed as the concatenation of the embeddings of the connected nodes,
 $   \vec{h}_{(u,v)} = \mathtt{Concat}(\vec{h}_u, \vec{h}_v)$.
The pre-trained \( \mathtt{GE} \) with learned weights \( \Theta_0 \) can be fine-tuned for downstream tasks.

\section{Proposed Approach}
In this section, we introduce \model, starting with a high-level overview of the framework, and then detail its core components.

\subsection{Overall Framework}
\begin{figure*}[t]
\centering
\includegraphics[width=0.9\linewidth]{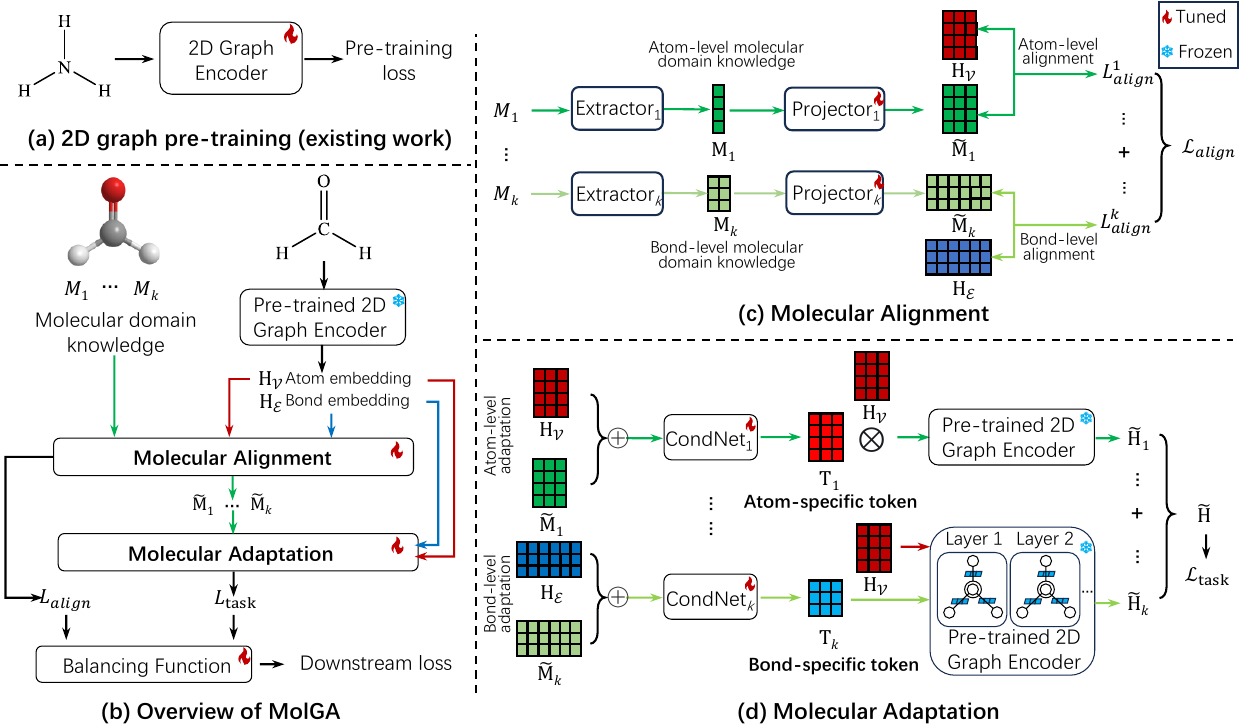}
\vspace{-2mm}
\caption{Overall framework of \model. Building upon an existing pre-trained 2D encoder, \model\ performs molecular alignment and adaptation for downstream tasks.}
\label{fig.framework}
\end{figure*}

We illustrate the overall framework of \model\ in Fig.~\ref{fig.framework}(b), which comprises two core components: molecular alignment and molecular adaptation. 
Given a pre-trained 2D graph encoder in Fig.~\ref{fig.framework}(a), \model\ first aligns the pre-trained topological representations with downstream molecular domain knowledge, as shown in Fig.~\ref{fig.framework}(c). For each type of molecular domain knowledge, \model\ trains a projector to map its representation into the pre-trained embedding space, and further employs a contrastive strategy to align the submolecular instances across their topological and molecular knowledge-based representations. 
Next, conditioned on the aligned representations, we employ conditional networks to generate instance-specific tokens, as depicted in Fig.~\ref{fig.framework}(d). These tokens incorporate fine-grained, instance-specific molecular domain knowledge into the adaptation process without updating the pre-trained 2D graph encoder, either by modifying the input features to the frozen encoder or by injecting them into the message-passing process.

\subsection{Molecular Alignment}

\stitle{2D topological encoding.}
We first feed the 2D version of the downstream molecular graph $G$ into the pre-trained 2D graph encoder $\mathtt{GE}$ to obtain the atom embedding matrix $\vec{H}_\bV$  and the bond embedding matrix $\vec{H}_\bE$ as introduced in Sect.~\ref{sec.preliminary}.

\stitle{Molecular domain knowledge extraction.}
For the $k$-th type of domain knowledge $M_k\in \mathcal{M}$, we employ a rule-based extractor $\mathtt{Extr}_k(\cdot)$, following prior work \citep{wang2022comenet}, to obtain the initial representations of $M_k$, as follows.
\begin{align}\label{eq.extract}
   \Vec{M}_k = \mathtt{Extr}_k(M_k).
\end{align}
The rule-based extractors are non-parametric and their details are presented in Appendix~\ref{app.parameters}.

\stitle{Molecular domain knowledge projection.}
Next, we map the initial representations of each type of molecular domain knowledge into the pre-trained atom embedding space by learnable projectors:
\begin{align}\label{eq.projector}
    \Vec{\Tilde{M}}_k=f_k(\Vec{M}_k;\phi_k),
\end{align}
where $f_k(\cdot)$ denotes the projector for the $k$-th type of knowledge, parameterized by learnable weights $\phi_k \in \Phi$. 
The output
$\Vec{\Tilde{M}}_k$ shares the same dimension as its corresponding atom embedding $\vec{H}_\bV$ or edge embedding $\vec{H}_\bE$, depending on whether $M_k$ is atom-level or bond-level knowledge. 
Let $\vec{\Tilde{m}_{i,k}}$ be the $i$-th row in $\Vec{\Tilde{M}}_k$, corresponding to the $k$-th type information for the $i$-th atom or bond. 

\stitle{Contrastive alignment.}
While the projection ensures that $\Vec{\Tilde{M}}_k$ and the instance representations ($\vec{H}_\bV$ or $\vec{H}_\bE$) share the same embedding dimension, they are not inherently aligned. Thus, we employ a contrastive alignment strategy to explicitly bridge the representations derived from molecular domain knowledge and topology.
Ideally, the 2D topological representation of an instance should implicitly reflect its chemical and physical properties \citep{pairault2023catenane,koppe2025coordination}. Hence, similar in spirit to multimodal alignment \citep{radford2021learning}, we adopt a contrastive learning strategy to align the two sources of representation for each instance. Specifically, we pull the 2D and domain-knowledge representations of the same instance closer in the latent space, while pushing apart those from different instances.
Formally, given a molecular graph $G$, the contrastive alignment loss for the $k$-th type of domain knowledge ($M_k$) is defined as:
\begin{equation}\label{eq:alignment_loss}\textstyle 
     L_{\text{align}}^{k}(\phi_k)= -\sum_{a\in G}\ln\frac{\exp(\text{sim}(\vec{h}_{a}, \vec{\Tilde{m}}_{a,k})/\tau)}{\sum_{b=1}^{S}\exp(\text{sim}(\vec{h}_{a}, \vec{\Tilde{m}}_{b,k})/\tau)},
\end{equation}
where $\vec{h}_a$ is the embedding of instance $a$. $S=|\bV|$ or $|\bE|$ is the number of instances in $G$, depending on the type of information. The temperature parameter $\tau$ is used to modulate the sharpness of the cosine similarity function $\mathrm{sim}(\cdot)$. 
Thus, the overall molecular alignment loss is:
\begin{align}\label{eq.alignment-loss}
\textstyle 
    \bL_\text{align}(\Phi)=\sum_{k=1}^K L_{\text{align}}^{k}(\phi_k).
\end{align}

\subsection{Molecular Adaptation}
\stitle{Instance-specific token generation.}
The aligned 2D topological and molecular knowledge representations are then used for downstream adaptation. Since submolecular instances, including atoms and bonds, manifest distinct topological, physical and chemical characteristics, it is advantageous to adapt to fine-grained, instance-level variations. To this end, for each type of domain knowledge, we employ a conditional network \citep{zhou2022conditional,yu2024node,yu2024non} that conditions on the fused 2D and molecular knowledge representations, and dynamically generates atom- or bond-specific tokens to guide the downstream instance encoding process in an instance-aware manner, thus effectively adapting the pre-trained 2D graph encoder to downstream tasks without updating the entire encoder.
Formally, for the $k$-th type of knowledge, we train a conditional network $\mathtt{CondNet}_k$ parameterized by $\gamma_k$, which generates a set of tokens as follows:
\begin{align}\label{eq:token_gen}
    \vec{T}_k = \mathtt{CondNet}_k(\mathrm{Concat}(\vec{H}, \vec{\Tilde{M}}_k); \gamma_k),
\end{align}
where $\vec{H}$ stands for the atom embeddings $\vec{H}_\bV$ or bond embeddings $\vec{H}_\bE$ depending on the type of knowledge. The conditional network acts as a hypernetwork \citep{ha2022hypernetworks}---a compact neural network, such as a multi-layer perceptron (MLP)---that generates instance-specific tokens for the primary network. $\vec{T}_k$ share the same dimension with the feature embedding $\vec{X}$. Each row of $\vec{T}_k$, denoted $\vec{t}_{i,k}$, is a unique adaptation token for the $i$-th instance.

\stitle{Instance adaptation.}
The generated tokens are then used to guide the adaptation of the pre-trained 2D graph encoder, enabling effective integration with molecular domain knowledge for downstream tasks. On one hand, for atom-level knowledge, atom-specific tokens modulate atom features as
\begin{align}\label{eq.atom-modification}
    \vec{\Tilde{X}}_k = \vec{X} \odot \vec{T}_k,
\end{align}
where $\vec{X}$ is the original atom feature matrix, $\vec{T}_k$ is the token matrix for the $k$-th type of knowledge (for atoms), and $\odot$ represents element-wise multiplication. The modified features $\vec{\Tilde{X}}_k$ are then fed into the pre-trained graph encoder, obtaining the output embedding $\vec{\Tilde{H}}_k$ through the frozen 2D graph encoder (Eq.~\ref{eq.gnn}).
On the other hand, for bond-level knowledge, we inject the bond-specific tokens into the message-passing process. Specifically, the $l$-th layer message passing is adapted as
\begin{align}\label{eq:structure_aggregation}
    \vec{h}^l_{v,k} = \mathtt{MP}\left(\vec{h}^{l-1}_{v,k}, \left\{\vec{t}_{(u,v),k} \cdot \vec{h}^{l-1}_{u,k} \mid u \in \mathcal{N}_v \right\}; \theta^l \right), \forall v \in \mathcal{V}.
\end{align}
We then aggregate the $K$ output embeddings conditioned on the $K$ types of molecular domain knowledge, as follows.
\begin{align}\textstyle
    \vec{\Tilde{H}}_\bV = \sum_{k=1}^K \vec{\Tilde{H}}_{\bV,k}.
\end{align}
Finally, we readout the representation of a molecular graph $G$ by sum pooling:
\begin{align}\textstyle
    \vec{H}_G=\mathtt{Readout}(\vec{\Tilde{H}}_\bV).
\end{align}

\stitle{Optimization.}
We focus on two popular downstream tasks, namely, molecular property prediction and molecular classification. Given a labeled dataset $\mathcal{D}_\text{down} = \{(G_1, y_1), (G_2, y_2), \dots\}$ for a downstream task, where \( G_i \) denotes a molecular graph and \( y_i \) is its associated label (i.e., a real-valued property or class label), we train a task head to infer the label of a given graph $G$. The task loss is denoted as  $\mathcal{L}_\text{task}(\Phi, \Gamma, \eta)$, where $\eta$ represents the learnable parameters in the task head, while $\Phi$ and $\Gamma$ represent the parameters of the projectors and conditional networks, respectively.
We integrate this task loss with the contrastive alignment loss (Eq.~\ref{eq.alignment-loss}) to form the final optimization objective:
\begin{align}\label{eq:downstream_loss}
    \mathcal{L}(\Phi, \Gamma, \eta, \beta) = \mathtt{Bal}(\mathcal{L}_\text{task}(\Phi, \Gamma, \eta), \mathcal{L}_\text{align}(\Phi);\beta),
\end{align}
where $\mathtt{Bal}(\cdot)$ is a balancing function parameterized by $\beta$. In our experiments, we adopt GradNorm~\citep{chen2018gradnorm} to adaptively balance the contributions of the two objectives.
During downstream adaptation, only $\Phi$ in the projectors,  $\Gamma$ in the conditional networks, $\eta$ in the task head and  $\beta$ in the balancing function are updated, whereas $\Theta$ in the pre-trained encoder are kept frozen. This parameter-efficient design ensures strong performance even under low-resource settings, where the downstream training set $\mathcal{D}$ contains only a few labeled examples.
We outline the algorithm of \model\ in Appendix~\ref{app.alg}.

\subsection{SE(3)-Invariance}
Molecular properties are invariant to the choice of global coordinate frame: applying a rigid rotation or translation to a molecule does not alter its intrinsic chemistry or the labels in typical property prediction tasks. Thus, when a downstream adaptation module leverages 3D conformations, its outputs should respect such symmetry. In what follows, we introduce $SE(3)$-invariance and prove that \model\ is invariant under any global rigid motion.

\begin{definition}[$SE(3)$-invariant representation \cite{wang2022comenet}]\label{def:se3-invariant}
Let $\mathbf{P}\in \mathcal{M}$ denote a molecular conformation, where each row $\vec{p}$ is the 3D coordinate of an atom. A global rigid motion is an element $g\in SE(3)$ acting as
\begin{equation}
g(\vec{p})=\mathbf{R}\vec{p}+\vec{t},\quad \mathbf{R}\in SO(3),\ \vec{t}\in\mathbb{R}^3,
\end{equation}
where $SO(3)$ denotes the 3D special orthogonal group. We overload $g(\mathbf{P})$ to apply the same transformation to all rows of $\mathbf{P}$. Thus, a mapping $f(G,\mathbf{P})$ is \emph{$SE(3)$-invariant} if
\begin{equation}\label{eq.se3-inv}
f(G,g(\mathbf{P})) = f(G,\mathbf{P}),\quad \forall g\in SE(3).
\end{equation}
\end{definition}

\begin{proposition}[\model\ is $SE(3)$-invariant]\label{prop:se3-invariant}
Let $\Psi(G,\mathbf{P})$ denote the deterministic forward mapping of \model\ downstream adaptation. $\Psi$ is $SE(3)$-invariant with respect to $\mathbf{P}$:
\begin{equation}
\Psi(G,g(\mathbf{P}))=\Psi(G,\mathbf{P}),\quad \forall g\in SE(3).
\label{eq.psi-inv}
\end{equation}
\end{proposition}

The proof of Proposition~\ref{prop:se3-invariant} is presented in Appendix~\ref{app:geo}.

\section{Experiments}\label{sec.exp}
In this section, we conduct experiments to evaluate \model\ and analyze the empirical results.

\subsection{Experimental Setup}\label{sec:expt:setup}
\stitle{Datasets.}
We utilize twelve benchmark datasets for evaluation, with \textit{BBBP} \citep{martins2012bayesian}, \textit{SIDER} \citep{kuhn2016sider}, \textit{ClinTox} \citep{gayvert2016data}, \textit{MUV} \citep{rohrer2009maximum} and \textit{BACE} \citep{wu2018moleculenet} for molecular classification task, \textit{QM8} \citep{ramakrishnan2015electronic}, \textit{QM9} \citep{ramakrishnan2014quantum}, \textit{ESOL} \citep{delaney2004esol}, \textit{Lipophilicity} \citep{gaulton2012chembl}, \textit{FreeSolv} \citep{mobley2014freesolv}, PCQM4Mv2\citep{hu2021ogb} and \textit{MD17-aspirin} \citep{chmiela2017machine} for molecular property prediction task. We summarize these datasets in Table~\ref{tab:dataset}.

\begin{table}[t]
\centering
\caption{Summary of the datasets.}
\vspace{-2mm}
\addtolength{\tabcolsep}{-1mm}
\label{tab:dataset}
\resizebox{1\columnwidth}{!}{ 
\begin{tabular}{lcrrrc}
\toprule
\textbf{Dataset} & \textbf{Task} & \textbf{Molecules} & \textbf{Avg. atoms} & \textbf{Avg. bonds} & \textbf{Molecular  knowledge} \\
\midrule
BBBP           & \multirow{5}{*}{\rotatebox{90}{Classification}}         & 2,039   & 24.1 & 26.0 & Chemical bonds \\
SIDER          &          & 1,427   & 33.6 & 35.4 & Chemical bonds \\
ClinTox        &          & 1,478   & 26.2 & 27.9 & Chemical bonds \\
MUV            &          & 93,087  & 24.2 & 26.3 & Chemical bonds \\
BACE           &          & 1,513   & 34.1 & 36.9 & Chemical bonds \\
\midrule
QM8            & \multirow{7}{*}{\rotatebox{90}{Property prediction}}     & 21,786  & 16.1 & 16.3 & 3D conformation \\
QM9            &     & 133,885 & 18.0 & 18.4 & 3D conformation \\
ESOL           &    & 1,128   & 13.3 & 13.7 & Chemical bonds \\
Lipophilicity  &     & 4,200   & 27.0 & 29.5 & Chemical bonds \\
FreeSolv       &    & 643     & 8.7  & 8.4  & Chemical bonds \\
PCQM4Mv2     &     & 3,746,619 & 14.1 & 14.6 & Chemical bonds \\
MD17-aspirin &     & 211,762 & 21   & 140.4 & 3D conform. \& atomic forces \\
\bottomrule
\end{tabular}}
\end{table}


\stitle{Baselines.}
We compare \model\ with four types of state-of-the-art methods:
(1) Supervised 2D graph encoders: \method{GCN} \citep{kipf2016semi} and \method{GAT} \citep{velivckovic2017graph} are trained directly on downstream labeled data in a fully supervised manner, without any pre-training or use of molecular domain knowledge.
(2) Supervised molecular graph encoders: \method{DimeNet++}~\citep{gasteiger2020fast}, \method{SphereNet}~\citep{liu2021spherical}, and \method{ComENet}~\citep{wang2022comenet} incorporate molecular domain knowledge into the message passing during downstream training, while \method{SE(3)-Transformer} performs self-attention directly on geometric tensor features. They are all end-to-end methods without pre-training.
(3) 2D graph pretraining method: \method{GraphCL}~\citep{you2020graph} and \method{GraphPrompt}~\citep{liu2023graphprompt} conduct unsupervised pre-training on unlabeled 2D graphs. They are later adapted to downstream tasks via fine-tuning or prompt learning, while keeping the pre-trained encoder frozen.
(4) Molecular graph pretraining methods: \method{GraphMVP}~\citep{liupre}, \method{GEM}~\citep{fang2022geometry}, \method{MoleBlend}~\citep{yu2023multimodal} and \method{Uni-Mol2}~\citep{ji2024uni} leverage molecular domain knowledge during both the pre-training and adaptation phases. They design self-supervised pre-training objectives grounded in molecular domain knowledge to learn molecular graph encoders, which are subsequently fine-tuned in a uniform manner across all instances for downstream tasks.
Additional implementation details for both the baselines and \model\ are detailed in Appendix~\ref{app.parameters}.

\begin{table*}[tbp] 
    \centering
    \footnotesize
    \caption{ROC-AUC (\%) evaluation of molecular classification.}
    \vspace{-2mm}
    \addtolength{\tabcolsep}{0.5mm}
    \label{table.graph-classification-moleculenet}
    \resizebox{0.9\linewidth}{!}{
    \begin{tabular}{l|cc|ccccc}
    \toprule
    {Methods} 
    & Pre-train & Adapt & BBBP & SIDER & ClinTox & MUV & BACE \\
    \midrule\midrule
    \method{GCN} & $\times$  & 2D
      & 66.23$\pm$13.83 
      & \underline{52.43}$\pm$\phantom{0}5.30 
      & 54.59$\pm$\phantom{0}5.33 
      & 52.48$\pm$\phantom{0}9.91 
      & 50.53$\pm$\phantom{0}6.20 \\
    \method{GAT} & $\times$  & 2D
      & 59.72$\pm$12.42 
      & 50.84$\pm$\phantom{0}5.50 
      & 52.91$\pm$\phantom{0}4.57 
      & 51.43$\pm$10.41 
      & 50.21$\pm$\phantom{0}5.55 \\
    \method{SE(3)-Transformer} & $\times$ & 2D+Mol
      & 67.02$\pm$\phantom{0}9.30
      & 51.75$\pm$\phantom{0}4.20
      & 52.11$\pm$\phantom{0}2.67
      & 54.88$\pm$17.10
      & 53.21$\pm$18.82 \\

    \midrule
    \method{DimeNet++} & $\times$ & 2D+Mol 
      & 66.41$\pm$10.43 
      & 50.85$\pm$\phantom{0}4.83 
      & 54.25$\pm$10.54 
      & 54.54$\pm$\phantom{0}7.66 
      & 54.06$\pm$13.00 \\
    \method{SphereNet} & $\times$ & 2D+Mol  
      & 65.54$\pm$10.51 
      & 50.26$\pm$\phantom{0}5.92 
      & 53.83$\pm$10.87 
      & 55.40$\pm$\phantom{0}5.79 
      & 54.43$\pm$13.41 \\
    \method{ComENet} & $\times$ & 2D+Mol  
      & 66.89$\pm$\phantom{0}9.94 
      & 50.52$\pm$\phantom{0}6.22 
      & \underline{55.89}$\pm$\phantom{0}9.09 
      & 56.69$\pm$\phantom{0}6.21 
      & \underline{54.57}$\pm$10.21 \\
    \midrule
    \method{GraphCL} & 2D & 2D 
      & 64.61$\pm$12.24 
      & 51.59$\pm$\phantom{0}4.45 
      & 55.58$\pm$\phantom{0}6.84 
      & 58.35$\pm$\phantom{0}9.03 
      & 52.20$\pm$\phantom{0}6.19 \\
    \method{GraphPrompt} & 2D & 2D   
      & \underline{68.88}$\pm$11.03 
      & 51.53$\pm$\phantom{0}4.20 
      & 55.67$\pm$\phantom{0}7.08 
      & \underline{59.87}$\pm$\phantom{0}6.27 
      & 52.43$\pm$\phantom{0}3.62 \\
    \midrule
    \method{\model} & 2D & 2D+Mol 
      & \textbf{71.97}$\pm$12.34 
      & \textbf{52.61}$\pm$\phantom{0}5.42 
      & \textbf{57.31}$\pm$\phantom{0}6.87 
      & \textbf{62.72}$\pm$\phantom{0}5.44 
      & \textbf{54.94}$\pm$\phantom{0}5.37 \\
    \midrule
    \rowcolor{gray!20}
    \multicolumn{8}{c}{Molecular graph pre-training methods with larger dataset and domain knowledge (for reference only)}\\
    \midrule
    \rowcolor{gray!20}
    \method{GraphMVP} & 2D+Mol & 2D+Mol   
      & 69.05$\pm$\phantom{0}8.00 
      & 50.82$\pm$\phantom{0}4.19 
      & 57.43$\pm$\phantom{0}6.12 
      & 59.59$\pm$\phantom{0}7.75 
      & 55.46$\pm$\phantom{0}6.10 \\
    \rowcolor{gray!20}
    \method{GEM} & 2D+Mol & 2D+Mol 
      & 71.78$\pm$\phantom{0}7.80 
      & 54.17$\pm$\phantom{0}6.30 
      & 62.32$\pm$\phantom{0}7.31 
      & 53.21$\pm$\phantom{0}8.03 
      & 58.07$\pm$\phantom{0}9.24 \\
    \rowcolor{gray!20}
    \method{MoleBlend} & 2D+Mol & 2D+Mol 
      & 71.45$\pm$\phantom{0}9.45 
      & 52.85$\pm$\phantom{0}4.25 
      & 60.56$\pm$\phantom{0}6.36 
      & 61.81$\pm$\phantom{0}7.96 
      & 57.17$\pm$\phantom{0}8.19 \\
    \rowcolor{gray!20}
    \method{Uni-Mol2} & 2D+Mol & 2D+Mol 
      & 72.16$\pm$\phantom{0}0.24
      & 53.85$\pm$\phantom{0}0.40
      & 61.61$\pm$\phantom{0}6.89
      & 61.40$\pm$\phantom{0}8.09
      & 56.07$\pm$\phantom{0}6.71 \\
    \midrule
    \end{tabular}}
    \parbox{0.88\linewidth}{\scriptsize ``$\times$'' indicates without pre-training; ``2D'' denotes the use of 2D topology; ``Mol'' signifies that molecular domain knowledge is leveraged. Among the comparable methods (excluding those for reference only), the best results are \textbf{bolded} and the second-best results are \underline{underlined}.}
\end{table*}
\begin{table*}[tbp] 
    \centering
    \scriptsize
    \addtolength{\tabcolsep}{-0.5mm}
    \caption{RMSE evaluation of molecular property prediction.}
    \vspace{-2mm}
    \label{table.graph-regression-moleculenet}
    \resizebox{0.9\linewidth}{!}{
    \footnotesize
    \begin{tabular}{l|cc|ccccccc}
    \toprule
    {Methods} 
    & Pre-train & Adapt & QM8 & QM9 & ESOL & Lipophilicity & FreeSolv & MD17-aspirin & PCQM4Mv2 \\
    \midrule\midrule
    \method{GCN} & $\times$ & 2D
      & 0.118{\tiny$\pm$0.003}
      & 1.400{\tiny$\pm$0.023}
      & 2.764{\tiny$\pm$0.122}
      & 1.466{\tiny$\pm$0.131}
      & 4.298{\tiny$\pm$0.143}
      & 4.872{\tiny$\pm$0.061}
      & 1.816{\tiny$\pm$0.075} \\
    \method{GAT} & $\times$ & 2D
      & 0.120{\tiny$\pm$0.003}
      & 1.160{\tiny$\pm$0.020}
      & 2.635{\tiny$\pm$0.116}
      & 1.445{\tiny$\pm$0.068}
      & 4.267{\tiny$\pm$0.203}
      & 4.958{\tiny$\pm$0.022}
      & 1.746{\tiny$\pm$0.029} \\
    \method{SE(3)-Transformer} & $\times$ & 2D+Mol
      & 0.085{\tiny$\pm$0.005}
      & \underline{0.912}{\tiny$\pm$0.670}
      & 2.592{\tiny$\pm$0.126}
      & 1.357{\tiny$\pm$0.112}
      & 3.937{\tiny$\pm$0.131}
      & 4.846{\tiny$\pm$0.020}
      & 1.650{\tiny$\pm$0.020} \\
    \midrule
    \method{DimeNet++} & $\times$ & 2D+Mol
      & 0.104{\tiny$\pm$0.004}
      & 1.034{\tiny$\pm$0.054}
      & 2.377{\tiny$\pm$0.078}
      & 1.379{\tiny$\pm$0.133}
      & 4.049{\tiny$\pm$0.213}
      & 4.781{\tiny$\pm$0.092}
      & 1.657{\tiny$\pm$0.108} \\
    \method{SphereNet} & $\times$ & 2D+Mol
      & 0.072{\tiny$\pm$0.007}
      & 0.928{\tiny$\pm$0.023}
      & 2.135{\tiny$\pm$0.151}
      & 1.335{\tiny$\pm$0.127}
      & \underline{3.900}{\tiny$\pm$0.138}
      & 4.772{\tiny$\pm$0.066}
      & 1.633{\tiny$\pm$0.051} \\
    \method{ComENet} & $\times$ & 2D+Mol
      & \underline{0.060}{\tiny$\pm$0.013}
      & 0.918{\tiny$\pm$0.042}
      & 2.317{\tiny$\pm$0.210}
      & 1.309{\tiny$\pm$0.048}
      & 4.049{\tiny$\pm$0.285}
      & \underline{4.755}{\tiny$\pm$0.081}
      & 1.581{\tiny$\pm$0.168} \\
    \midrule
    \method{GraphCL} & 2D & 2D
      & 0.103{\tiny$\pm$0.007}
      & 1.018{\tiny$\pm$0.011}
      & \underline{2.038}{\tiny$\pm$0.047}
      & 1.249{\tiny$\pm$0.014}
      & 4.612{\tiny$\pm$0.133}
      & 4.900{\tiny$\pm$0.145}
      & 1.460{\tiny$\pm$0.029} \\
    \method{GraphPrompt} & 2D & 2D
      & 0.100{\tiny$\pm$0.096}
      & 0.976{\tiny$\pm$0.022}
      & 2.052{\tiny$\pm$0.045}
      & \underline{1.242}{\tiny$\pm$0.017}
      & 4.517{\tiny$\pm$0.129}
      & 4.843{\tiny$\pm$0.068}
      & \underline{1.424{\tiny$\pm$0.071}} \\
    \midrule
    \model & 2D & 2D+Mol
      & \textbf{0.056}{\tiny$\pm$0.001}
      & \textbf{0.856}{\tiny$\pm$0.026}
      & \textbf{1.973}{\tiny$\pm$0.043}
      & \textbf{1.187}{\tiny$\pm$0.015}
      & \textbf{3.602}{\tiny$\pm$0.131}
      & \textbf{4.755}{\tiny$\pm$0.049}
      & \textbf{1.287}{\tiny$\pm$0.016} \\
    \midrule
    \rowcolor{gray!20}
    \multicolumn{10}{c}{Molecular graph pre-training methods with larger dataset and domain knowledge (for reference only)}\\
    \midrule
    \rowcolor{gray!20}
    \method{GraphMVP} & 2D+Mol & 2D+Mol
      & 0.032{\tiny$\pm$0.004}
      & 0.885{\tiny$\pm$0.023}
      & 2.257{\tiny$\pm$0.043}
      & 0.945{\tiny$\pm$0.014}
      & 2.767{\tiny$\pm$0.122}
      & 4.782{\tiny$\pm$0.033}
      & 1.232{\tiny$\pm$0.018} \\
    \rowcolor{gray!20}
    \method{GEM} & 2D+Mol & 2D+Mol
      & 0.028{\tiny$\pm$0.011}
      & 0.652{\tiny$\pm$0.003}
      & 1.175{\tiny$\pm$0.020}
      & 0.840{\tiny$\pm$0.220}
      & 1.957{\tiny$\pm$0.073}
      & 4.761{\tiny$\pm$0.014}
      & 1.205{\tiny$\pm$0.013} \\
    \rowcolor{gray!20}
    \method{MoleBlend} & 2D+Mol & 2D+Mol
      & 0.030{\tiny$\pm$0.013}
      & 0.590{\tiny$\pm$0.009}
      & 1.496{\tiny$\pm$0.032}
      & 0.774{\tiny$\pm$0.012}
      & 1.851{\tiny$\pm$0.068}
      & 4.708{\tiny$\pm$0.082}
      & 1.176{\tiny$\pm$0.010} \\
    \rowcolor{gray!20}
    \method{Uni-Mol2} & 2D+Mol & 2D+Mol
      & 0.033{\tiny$\pm$0.016}
      & 0.538{\tiny$\pm$0.016}
      & 1.261{\tiny$\pm$0.021}
      & 0.731{\tiny$\pm$0.048}
      & 1.792{\tiny$\pm$0.135}
      & 4.640{\tiny$\pm$0.024}
      & 1.142{\tiny$\pm$0.013} \\
    \midrule
    \end{tabular}}
\end{table*}

\stitle{Pre-training setting.}
For all baselines and \model, we randomly sample 20\% of molecules from the \textit{QM9} dataset~\citep{ramakrishnan2014quantum}, resulting in a subset of 25,000 molecules. 
Specifically, for 2D graph pre-training methods and \model, only the 2D molecular graphs—without any molecular domain knowledge—are used for pre-training. 
To further evaluate the effectiveness of \model, we also compare it with molecular graph pre-training methods, which leverage both the 2D graph structure and the associated molecular domain knowledge during pre-training and adaptation.
We leverage their officially released pre-trained molecular graph encoders, which are trained on significantly larger datasets: \method{GraphMVP} is pre-trained on GEOM \citep{axelrod2022geom} with 50,000 molecules, \method{GEM} on ZINC15 \citep{sterling2015zinc} with 2,000,000 molecules, and \method{MoleBlend} on PCQM4Mv2 \citep{hu2021ogb} with 3,370,000 molecules. Since molecular graph pre-training methods utilize substantially more data and incorporate molecular domain knowledge during pre-training, they are not directly comparable to \model.

\begin{figure*}[t]
\centering
\includegraphics[width=0.95\linewidth]{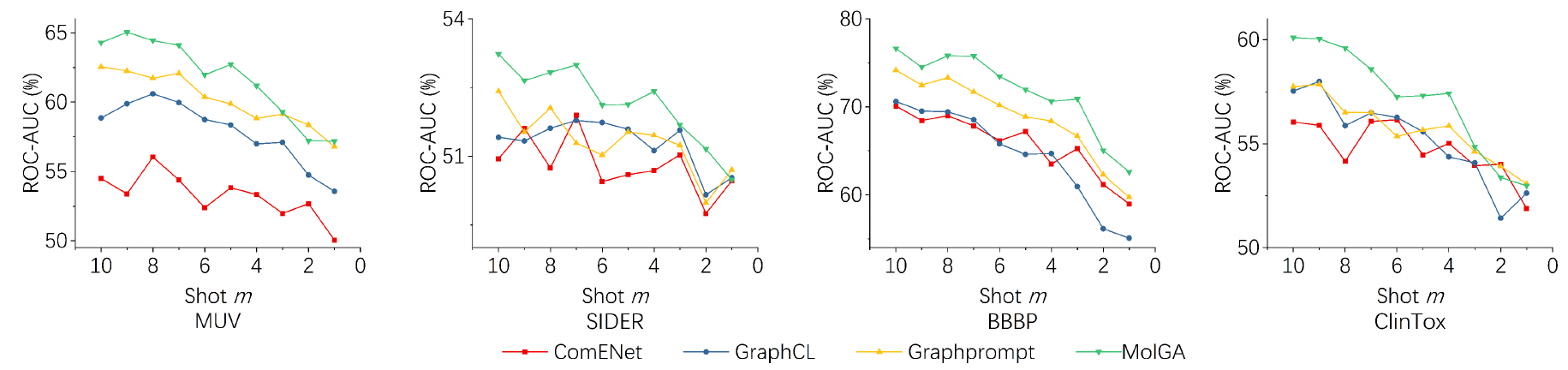}
\vspace{-4mm}
\caption{Impact of labeled data size (number of shots) on molecular classification.}
\label{fig.fewshot}
\end{figure*}

\begin{figure}[t]
\centering
\includegraphics[width=1\linewidth]{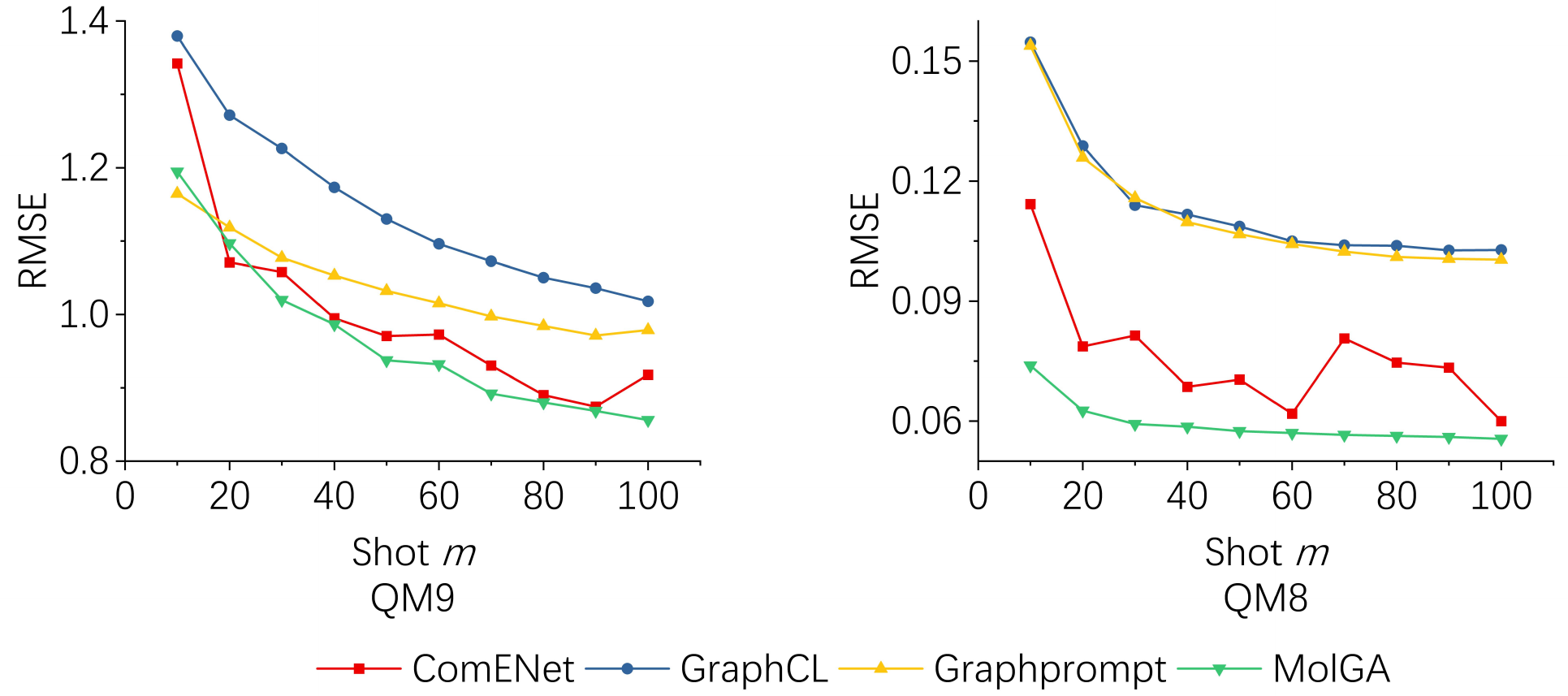}
\vspace{-4mm}
\caption{Impact of labeled data size  on molecular property prediction.}
\label{fig.fewshotqm9}
\end{figure}

\begin{table*}[tbp] 
    \centering
    \small
    \addtolength{\tabcolsep}{0mm}
    \caption{Ablation study on the effects of different components. }
    \vspace{-2mm}
    \label{table.ablation-components}
    \resizebox{0.9\linewidth}{!}{
    \begin{tabular}{l|ccc|cccc|cccc}
    \toprule
    \multirow{2}*{Methods}     & \multirow{2}*{AL} & \multicolumn{2}{c|}{CN}
    & \multicolumn{4}{c|}{Molecular classification} 
    & \multicolumn{4}{c}{Molecular property prediction} \\
    & &2D &Mol
    & BBBP & ClinTox & MUV & BACE 
    & QM8 & ESOL & Lipophilicity & FreeSolv \\
    \midrule\midrule
    \method{Variant 1} 
        & $\times$ & $\times$ & $\times$ 
        & 65.29{\tiny$\pm$12.04} & 55.88{\tiny$\pm$7.35} & 62.61{\tiny$\pm$4.77} & 52.61{\tiny$\pm$3.33}
        & 0.070{\tiny$\pm$0.001} & 2.035{\tiny$\pm$0.075} & 1.211{\tiny$\pm$0.023} & 3.710{\tiny$\pm$0.122} \\
    \method{Variant 2}
        & $\times$ & $\checkmark$ & $\times$  
        & 70.16{\tiny$\pm$11.59} & 56.66{\tiny$\pm$6.83} & 60.47{\tiny$\pm$6.48} & 53.45{\tiny$\pm$5.40}
        & 0.057{\tiny$\pm$0.001} & 2.063{\tiny$\pm$0.042} & 1.199{\tiny$\pm$0.015} & 3.716{\tiny$\pm$0.124} \\
    \method{Variant 3}
        & $\times$ & $\times$  & $\checkmark$ 
        & 69.87{\tiny$\pm$13.00} & 54.83{\tiny$\pm$6.69} & 59.05{\tiny$\pm$5.89} & 49.96{\tiny$\pm$2.61}
        & 0.057{\tiny$\pm$0.001} & 2.073{\tiny$\pm$0.057} & 1.200{\tiny$\pm$0.015} & 3.668{\tiny$\pm$0.097} \\
    \method{Variant 4}
        & $\times$ & $\checkmark$ & $\checkmark$ 
        & 71.10{\tiny$\pm$10.71} & 56.75{\tiny$\pm$6.78} & 61.40{\tiny$\pm$6.33} & 53.91{\tiny$\pm$5.98}
        & 0.059{\tiny$\pm$0.004} & 2.018{\tiny$\pm$0.055} & 1.186{\tiny$\pm$0.010} & 3.624{\tiny$\pm$0.208} \\
    \method{Variant 5} 
        & $\checkmark$ & $\times$ & $\times $
        & 71.53{\tiny$\pm$0.11} & 57.13{\tiny$\pm$6.02} & 60.51{\tiny$\pm$7.11} & 54.11{\tiny$\pm$5.86}
        & 0.059{\tiny$\pm$0.001} & 1.980{\tiny$\pm$0.042} & 1.193{\tiny$\pm$0.011} & 3.610{\tiny$\pm$0.144} \\
    \model 
        & $\checkmark$ & $\checkmark$ & $\checkmark$ 
        & \textbf{71.97}{\tiny$\pm$12.34} & \textbf{57.31}{\tiny$\pm$6.87} & \textbf{62.72}{\tiny$\pm$5.44} & \textbf{54.94}{\tiny$\pm$5.37}
        & \textbf{0.056}{\tiny$\pm$0.001} & \textbf{1.973}{\tiny$\pm$0.043} & \textbf{1.187}{\tiny$\pm$0.015} & \textbf{3.602}{\tiny$\pm$0.131} \\
    \bottomrule
    \end{tabular}}
    \parbox{0.88\linewidth}{\scriptsize 
``AL'' is short for molecular alignment; ``CN'' is short for conditional networks; ``2D'' denotes conditioning on pre-trained 2D topological knowledge; ``Mol'' stands for conditioning on molecular domain knowledge.
}
\end{table*}

\stitle{Adaptation setting.}
We evaluate \model\ on both molecular classification and molecular property prediction tasks.
For molecular classification, we follow an \(m\)-shot classification setting: for each class, \(m\) labeled instances are randomly selected for training, and the remaining instances are used for testing. In our main experiments, we set \(m = 5\), which corresponds to less than 1\% of the dataset.
For molecular property prediction, we randomly sample 100 molecules for downstream adaptation training, 100 for validation, and use the remaining molecules for testing. For \textit{QM9}, which is used for pre-training, the samples are drawn from the remaining 80\% of data not used during pre-training. For other datasets, sampling is performed from the full dataset.

\stitle{Evaluation.} 
For molecular classification, we follow a previous study \citep{wu2018moleculenet}, employing the ROC-AUC \citep{bradley1997use} as the evaluation metric. We construct 100 independent $m$-shot tasks via repeated sampling, and evaluate each task using five distinct random seeds, resulting in a total of 500 results.
For molecular property prediction, we adopt RMSE \citep{hodson2022root} as the evaluation metric, following prior work \citep{ruddigkeit2012enumeration} and assess with 5 different random seeds.
For both molecular classification and property prediction, we report the mean and standard deviation across all runs.

\subsection{Performance Evaluation}
We first evaluate \model\ on both molecular classification and property prediction tasks. Then we assess performance with various amounts of labeled data for downstream adaptation.

\stitle{Molecular classification and property prediction.} 
We present the results of molecular classification in Table~\ref{table.graph-classification-moleculenet} and molecular property prediction in Table~\ref{table.graph-regression-moleculenet}. We observe that: (1) \model\ consistently outperforms other baselines, demonstrating the effectiveness of incorporating molecular domain knowledge during the adaptation phase of 2D pre-trained graph encoders; (2) Since molecular graph pre-training methods utilize a  \textit{significantly larger ($2\sim 134\times$) amount of data} for pre-training and are thus not directly comparable, their results are reported only for reference. Nevertheless, \model\ still achieves competitive performance relative to them.

\begin{table*}[tbp] 
    \centering
    \footnotesize
    \caption{Evaluation of \model\ with different 2D graph pre-training methods.}
    \vspace{-2mm}
    \label{table.pretrain-method}
    \resizebox{0.9\linewidth}{!}{
    \begin{tabular}{l|cccc|cccc}
    \toprule
    \multirow{2}{*}{Pre-training Method}
      & \multicolumn{4}{c|}{Molecular classification} 
      & \multicolumn{4}{c}{Molecular property prediction } \\
      & SIDER & ClinTox & MUV & BACE 
      & QM8 & ESOL & Lipophilicity & FreeSolv \\
    \midrule\midrule
    JOAOv2  
      & 52.61$\pm$5.42
      & 57.31$\pm$6.87
      & 62.72$\pm$5.44
      & 54.94$\pm$5.37
      & 0.056{\tiny$\pm$0.001} 
      & 1.878{\tiny$\pm$0.042} 
      & 1.187{\tiny$\pm$0.015} 
      & 3.622{\tiny$\pm$0.131} \\
    GraphCL 
      & 51.61$\pm$5.33 
      & 57.74$\pm$6.23
      & 60.81$\pm$6.42
      & 54.17$\pm$5.71
      & 0.059{\tiny$\pm$0.001} 
      & 1.950{\tiny$\pm$0.044} 
      & 1.189{\tiny$\pm$0.015} 
      & 3.530{\tiny$\pm$0.092} \\
    \bottomrule
    \end{tabular}}
\end{table*}

\stitle{Impact of labeled data size.}
To assess the impact of labeled data size, we evaluate \model\ and several strong baselines on molecular classification tasks under varying numbers of shots, as shown in Fig.~\ref{fig.fewshot}. We make the following  observations.
(1) \model\ generally surpasses all competing baselines from 1 to 10 shots, demonstrating its robustness.
(2) As the number of labeled instances increases (e.g., $m > 5$), performance typically improves for all models. However, \model\ exhibits a steeper performance gain compared to other methods.
(3) In low-shot scenarios (e.g., $m \leq 5$), \model\ remains highly competitive, frequently achieving the best or near-best results.

We further examine how performance scales with the amount of labeled data in molecular property prediction. As shown in Fig.~\ref{fig.fewshotqm9}, all methods benefit from increasing the number of shots, yet \model\ consistently maintains a clear advantage, achieving the lowest RMSE across all label budgets, indicating strong transferability to regression tasks.

\subsection{Ablation Study}\label{sec.exp.ablation}

To understand the effect of integration with molecular domain knowledge, molecular alignment and molecular adaptation, we compare \model\ with five ablated variants and report the results in Table~\ref{table.ablation-components}. We observe that:
(1) Integrating molecular domain knowledge benefits the adaptation of pre-trained 2D graph encoders. This is evidenced by Variant 4 outperforming Variant 2. However, molecular knowledge alone is insufficient, as Variant 4 also surpasses Variant 3, suggesting the necessity of effective integration of both 2D topological and molecular domain knowledge.
(2) Molecular alignment effectively bridges the representational gap between pre-trained 2D topological embeddings and associated molecular knowledge, as Variant 5 outperforms Variant 1, and \model\ further outperforms Variant 4.  This highlights the advantage of aligning these two modalities.
(3) Conditional networks are beneficial for both molecular classification and molecular property prediction, as \model\ surpasses Variant 5, and Variant 2 outperforms Variant 1. This demonstrates that modeling instance-specific characteristics—rather than applying a uniform task head—is crucial for fully leveraging both 2D topological structures and molecular domain knowledge.


\subsection{Impact of Hyperparameters}\label{sec.exp.hyperparameters}
\begin{figure}[t]
\centering
\includegraphics[width=1\linewidth]{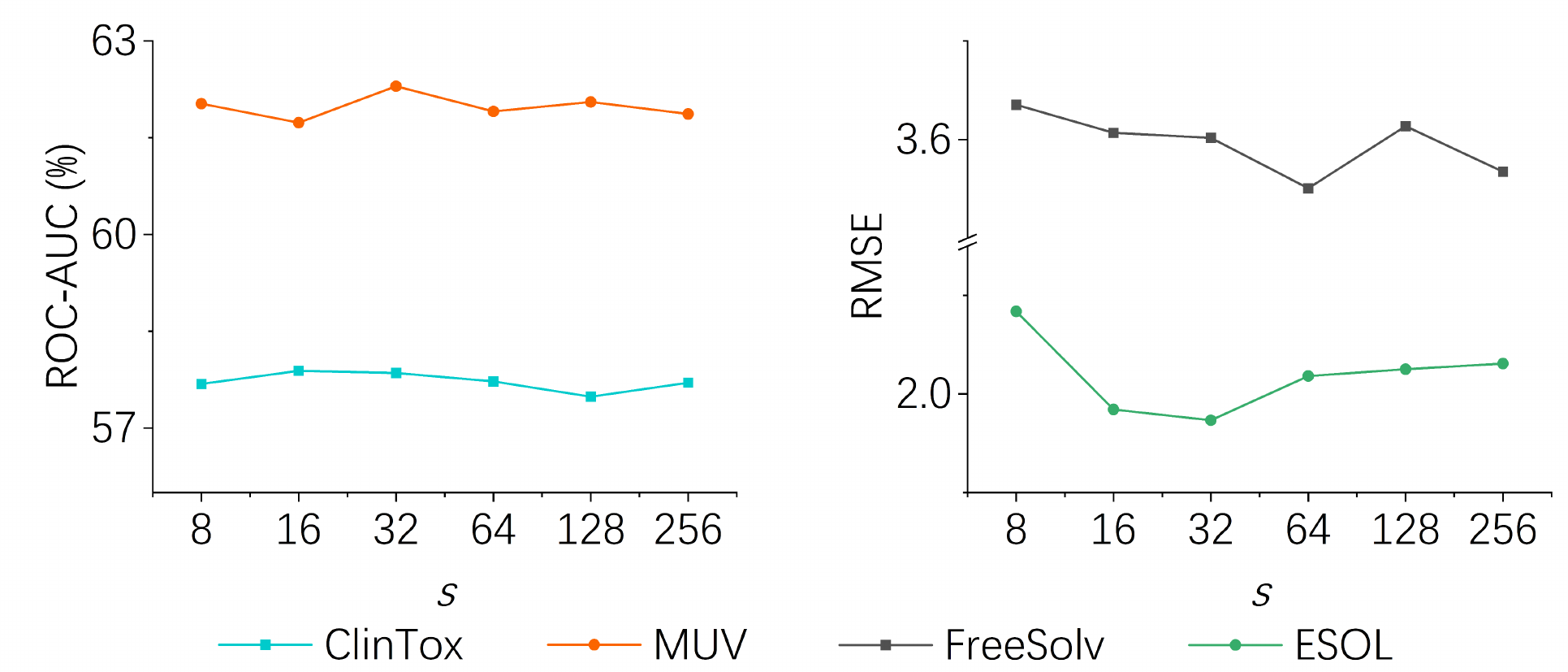}
\vspace{-4mm}
\caption{Impact of hidden dimension $s$ in the conditional networks.}
\label{fig.hiddendim}
\end{figure}
In our experiments, the conditional network is implemented as a two-layer MLP. To investigate the impact of its design, we vary the hidden dimension  \(s\) of the conditional network and report the results in Fig.~\ref{fig.hiddendim}. We observe that \(s = 32\) generally achieves robust performance on both molecular classification and molecular property prediction tasks, thus we set \(s = 32\) in our experiments.

\subsection{Impact of Pre-training Methods} \label{sec.pre-training-flexibility}
To assess the impact of different 2D graph pre-training methods, we employ two representative approaches, GraphCL and JOAOv2 \citep{you2021joao}, and report the results in Table~\ref{table.pretrain-method}. We observe that \model\ consistently outperforms comparable baselines across all pre-training methods, demonstrating its flexibility and general applicability. Moreover, using JOAOv2\ as the pre-training method generally yields better performance than GraphCL, thus we adopt JOAOv2 in our main experiments.


\begin{table}[tbp]
\centering
\small
\caption{Comparison of the number of tunable parameters during the downstream adaptation stage.}
\vspace{-2mm}
\label{table.parameters-num}
\begin{tabular}{l|r}
\toprule
Methods & No. parameters \\\midrule\midrule
\method{GCN}        & 33,537 \\
\method{GAT}        & 17,921 \\
\midrule
\method{DimeNet++}  & 1,919,750 \\
\method{SphereNet}  & 1,916,870 \\
\method{ComENet}    & 996,236 \\
\midrule
\method{GraphCL}    & 128 \\
\method{GraphPrompt}& 64 \\
\midrule
\method{\model}     & 13,824 \\\bottomrule
\end{tabular}
\end{table}

\subsection{Parameter Efficiency}\label{app.exp.effciency}
Finally, we evaluate the parameter efficiency of \model\ by comparing it with representative baselines. We report the number of parameters that need to be tuned during the downstream adaptation phase, as shown in Table~\ref{table.parameters-num}. We draw the following observations:
(1) Supervised learning methods, including \method{GCN}, \method{GAT}, \method{DimeNet++}, \method{SphereNet}, and \method{ComENet}, are trained end-to-end, requiring all model parameters to be updated during downstream training. Thus, these models exhibit poor parameter efficiency. Moreover, compared to \method{GCN} and \method{GAT}, which only utilize 2D topological structures, the others further incorporate molecular domain knowledge, resulting in a significantly larger number of trainable parameters.
(2) 2D graph pre-training-based methods, \method{GraphCL} and \method{GraphPrompt}, freeze the pre-trained 2D graph encoders during downstream adaptation, only tune a task head, significantly improving parameter efficiency. 
(3) \model\ requires substantially fewer tunable parameters than the supervised baselines. Although \model\ introduces additional lightweight modules (projectors and conditional networks) and therefore tunes slightly more parameters than 2D pre-training methods, the increase is modest and does not pose a practical bottleneck.

\section{Conclusions}

In this paper, we studied the adaptation of pre-trained 2D graph encoder to downstream molecular applications by flexibly integrating diverse molecular domain knowledge. Our proposed method, \model, employs a molecular alignment strategy to bridge the gap between the pre-trained 2D topological representation and downstream molecular domain knowledge. Furthermore, we propose a conditional molecular adaptation mechanism that modulates the pre-trained encoder conditioned on aligned topological and molecular knowledge. Extensive experiments on eleven benchmarks demonstrate the effectiveness of \model.




\begin{acks}
This research is supported by the Lee Kong Chian Fellowship awarded to Yuan Fang by Singapore Management University. We thank Mingjiang Zhang for valuable advice on the chemistry-related aspects of this work and assistance with figure preparation.
\end{acks}

\appendix
\section*{Appendices}

\begin{algorithm}[!b]
\small
\caption{\textsc{Downstream adaptation for \model}}
\label{alg.molga}
\begin{algorithmic}[1]
    \Require Pre-trained 2D graph encoder $\mathtt{GE}$ with frozen parameters $\Theta_0$, labeled data $\bD$ for downstream adaptation;
    \Ensure Optimized parameters $\Gamma$ in conditional networks, $\Phi$ in projectors, $\eta$ in task head, and $\beta$ in balancing function.
    \State  $\Phi,\Gamma,\eta,\beta \leftarrow$ initialization
    \While{not converged}
                \State $(\vec{H}_\bV,\vec{H}_\bE)\leftarrow \mathtt{GE}(\Vec{X},G;\Theta_0)$
                \State \slash* Molecular alignment *\slash
                \For{$k=1$ to $K$}
                    \State \slash* Molecular domain knowledge extraction by Eq.~\ref{eq.extract} *\slash
                    \State $\Vec{M}_k \leftarrow \mathtt{Extr}_k(M_k)$
                    \State \slash* Molecular domain knowledge projection by Eq.~\ref{eq.projector} *\slash
                    \State $\Vec{\Tilde{M}}_k \leftarrow f_k(\Vec{M}_k;\phi_k)$
                \EndFor
                \State \slash* Contrastive alignment loss *\slash
                \For{$k=1$ to $K$}
                    \State Calculate  $L_{\text{align}}^{k}(\phi_k)$ by Eq.~\ref{eq:alignment_loss}
                \EndFor
                \State $\bL_\text{align} \leftarrow \sum_{k=1}^K L_{\text{align}}^{k}$ (Eq.~\ref{eq.alignment-loss})
                \State \slash* Molecular adaptation *\slash
                \State \slash* Molecular token generation by Eq.~\ref{eq:token_gen} *\slash
                \For{$k=1$ to $K$}
                    \State $\vec{T}_k \leftarrow \mathtt{CondNet}_k\!\big(\mathrm{Concat}(\vec{H},\Vec{\Tilde{M}}_k);\gamma_k\big)$
                \EndFor
                \For{$k=1$ to $K$}
                    \If{$M_k$ is atom-level}
                        \State $\vec{\Tilde{X}}_k \leftarrow \vec{X}\odot \vec{T}_k$ (Eq.~\ref{eq.atom-modification})
                        \State $\vec{\Tilde{H}}_{\bV,k} \leftarrow \mathtt{GE}\big(\vec{\Tilde{X}}_k,G;\Theta_0\big)$ 
                    \ElsIf{$M_k$ is bond-level}
                        \For{Each layer $l$ in $\mathtt{GE}$}            
                            \State \parbox[t]{0.7\linewidth}{%
                            $\vec{h}^l_{v,k} = \mathtt{MP}\left(\vec{h}^{l-1}_{v,k}, \left\{\vec{t}_{(u,v),k} \cdot \vec{h}^{l-1}_{u,k} \mid u \in \mathcal{N}_v \right\}; \theta^l \right),  \forall v \in \mathcal{V}$ (Eq.~\ref{eq:structure_aggregation})}
                        \EndFor
                    \EndIf
                \EndFor
                \State $\vec{\Tilde{H}}_\bV \leftarrow \sum_{k=1}^K \vec{\Tilde{H}}_{\bV,k}$;\quad $\vec{H}_G \leftarrow \mathtt{Readout}(\vec{\Tilde{H}}_\bV)$
                \State \slash* Task loss *\slash
                \State $\hat{y}\leftarrow \mathtt{TaskHead}(\vec{H}_G;\eta)$;\quad $\mathcal{L}_\text{task}\leftarrow L(\hat{y},y)$
            \State Calculate $\mathcal{L}(\Phi, \Gamma, \eta, \beta)$ by Eq.~\ref{eq:downstream_loss}
            \State Update $\Phi, \Gamma, \eta, \beta$ by backpropagating  $\mathcal{L}(\Phi, \Gamma, \eta, \beta)$
    \EndWhile
    \State \Return $\{\Phi,\Gamma,\eta,\beta\}$
\end{algorithmic}
\end{algorithm}

\section{Algorithm}\label{app.alg}
We detail the key steps for downstream tuning of \model\ in Algorithm~\ref{alg.molga}. First, we extract molecular domain knowledge and project them into the topological embedding space (lines 5--9). Then, in lines 10--11, we calculate contrastive alignment loss. In lines 14--17, we employ conditional networks to generate molecular tokens, which are then used to modify molecular graphs in lines 18--24. Specifically, for atom-associated molecular domain knowledge, tokens modify the input atom features before being fed into the frozen 2D graph encoder (lines 19--21), while for bond-associated molecular domain knowledge, generated tokens are injected into the message-passing process (lines 22--24). Then we read out the molecular embedding and calculate task loss (lines 25--28). Finally, we calculate and backpropagate the overall loss and optimize $\Phi,\Gamma,\eta,\beta$ (lines 29--30).

\section{SE(3)-Invariance}\label{app:geo}

We first present Lemma~\ref{the.lemma} below, which shows the $SE(3)$-invariance of the extracted and projected geometric knowledge. 

\begin{lemma}[Invariance of geometric knowledge]\label{the.lemma}
Let $\mathbf{P}\in\mathbb{R}^{|\bV|\times 3}$ be a molecular conformation and let $g\in SE(3)$ be any global rigid motion.
Define the extracted geometric representation
\begin{equation}
\Vec{M}_{\mathbf{P}} = \mathtt{Extr}_{\mathbf{P}}(\mathbf{P}),
\end{equation}
and the projected representation (Eq.~\ref{eq.projector})
\begin{equation}
\Vec{\Tilde{M}}_{\mathbf{P}} = f_{\mathbf{P}}(\Vec{M}_{\mathbf{P}};\phi_{\mathbf{P}}).
\end{equation}
$f_{\mathbf{P}}$ is $SE(3)$-invariant, i.e.,
\begin{equation}
\Vec{\Tilde{M}}_{g(\mathbf{P})} = f_{\mathbf{P}}(\mathtt{Extr}_{\mathbf{P}}(g(\mathbf{P}));\phi_{\mathbf{P}})=\Vec{\Tilde{M}}_{\mathbf{P}},\quad \forall g\in SE(3).
\label{eq:assump-inv}
\end{equation}
\end{lemma}

\begin{proof}
$\mathtt{Extr}_{\mathbf{P}}$ is constructed from relative geometric quantities and is invariant to any global rotation/translation. Hence, for all $g\in SE(3)$,
\begin{equation}
\mathtt{Extr}_{\mathbf{P}}(g(\mathbf{P}))=\mathtt{Extr}_{\mathbf{P}}(\mathbf{P}).
\end{equation}
The projector $f_{\mathbf{P}}(\cdot;\phi_{\mathbf{P}})$ in Eq.~\ref{eq.projector} operates only on the extracted features in latent space
and does not access absolute coordinates. Therefore, applying $f_{\mathbf{P}}$ preserves invariance:
\begin{equation}
\Vec{\Tilde{M}}_{g(\mathbf{P})}
= f_{\mathbf{P}}(\mathtt{Extr}_{\mathbf{P}}(g(\mathbf{P}));\phi_{\mathbf{P}})
= f_{\mathbf{P}}(\mathtt{Extr}_{\mathbf{P}}(\mathbf{P});\phi_{\mathbf{P}})
= \Vec{\Tilde{M}}_{\mathbf{P}}.
\end{equation}
\end{proof}

Building on Lemma~\ref{the.lemma}, we derive Proposition~\ref{prop:se3-invariant}.

\begin{proof}\label{app:geo:proofprop1}
Recall that $\Psi$ denotes the deterministic forward mapping of \model.
The frozen 2D encoder produces node representations
$\vec{H}=\mathtt{GE}(\vec{X},G;\Theta_0)$ (Eq.~\ref{eq.gnn}), which depend only on $(G,\vec{X})$ and are therefore independent of $\mathbf{P}$.
By Lemma~1, the geometry-dependent knowledge used by \model\ is invariant under any rigid motion:
\begin{equation}
f_{\mathbf{P}}\!\left(\mathtt{Extr}_{\mathbf{P}}(g(\mathbf{P}));\phi_{\mathbf{P}}\right)
=
f_{\mathbf{P}}\!\left(\mathtt{Extr}_{\mathbf{P}}(\mathbf{P});\phi_{\mathbf{P}}\right),
\quad \forall g\in SE(3).
\end{equation}
Hence, all inputs to the token generator $\mathtt{CondNet}$ in Eq.~\ref{eq:token_gen} are $SE(3)$-invariant, which implies that the
generated tokens are also invariant:
\begin{equation}
\vec{T}_{\mathbf{P}}(G,g(\mathbf{P}))=\vec{T}_{\mathbf{P}}(G,\mathbf{P}),\quad \forall g\in SE(3).
\end{equation}
Finally, token injection in Eq.~\ref{eq.atom-modification} (feature-wise modulation) and Eq.~\ref{eq:structure_aggregation} (multiplicative gating on latent messages) is coordinate-free: these operations act only on latent features or messages and do not access $\mathbf{P}$ explicitly. Therefore, injecting $SE(3)$-invariant tokens into the geometry-free backbone produces identical adapted representations under $\mathbf{P}$ and $g(\mathbf{P})$.
Combining the above yields $\Psi(G,g(\mathbf{P}))=\Psi(G,\mathbf{P})$ for all $g\in SE(3)$,
which proves Proposition~1.
\end{proof}

\section{Implementation Details} \label{app.parameters}

\stitle{Details of baselines.}
For all open-source baselines, we leverage the officially provided code. Each method is tuned based on the settings recommended in their respective literature to achieve optimal performance.

For both GCN\citep{kipf2016semi} and GAT\citep{velivckovic2017graph}, we employ a 3-layer architecture, and set the hidden dimension to 64.

For GraphCL \citep{you2020graph}, a 3-layer GCN is also employed as its base model, with the hidden dimension set to 64. Specifically, we select edge dropping as the augmentation, with a default augmentation ratio of 0.2.

For GraphPrompt \citep{liu2023graphprompt}, a 3-layer GCN is used as the base model for all datasets, with the hidden dimension set to 64.

For DimeNet++ \citep{gasteiger2020fast}, we use the default settings provided in the original implementation: 4 interaction layers, hidden dimension of 128, and a cutoff radius of 5.0.

For SphereNet \citep{liu2021spherical}, we adopt the default configuration: 4 layers with a hidden dimension of 128 and cutoff radius of 5.0. The basis sizes are kept as default with 7 spherical harmonics and 6 radial basis functions. Output block settings are also maintained as default, with no additional hyperparameter tuning.

For ComENet \citep{wang2022comenet}, we employ the composite-energy message passing scheme to efficiently encode complete 3D information within 1-hop neighborhoods. Following the DIG framework defaults, we configure the model with 4 layers, a hidden dimension of 256, middle dimension of 64, and a cutoff radius of 8.0. The basis sizes are set to 3 spherical harmonics and 2 radial basis functions, and output layers remain unchanged.

For GraphMVP \citep{liupre}, we utilize a 5-layer GIN as the 2D encoder with a hidden dimension of 300. For the molecular graph encoder, we adopt a 6-layer SchNet \citep{schutt2017schnet} with a hidden dimension of 128.

For GEM \citep{fang2022geometry}, we employ GeoGNN \citep{fang2022geometry} as the molecular graph encoder, consisting of 8 blocks with a hidden dimension of 32.

For MoleBlend \citep{yu2023multimodal}, we use a unified Transformer backbone with 12 layers and 32 attention heads, where the hidden dimension is set to 768.

For Uni-Mol2 \citep{ji2024uni}, we adopt the 1.1B-scale two-track Transformer backbone with 64 blocks, a hidden dimension of 1536, and 96 attention heads; the pair track uses a 512-dimensional pair embedding with 64-dimensional pair hidden states and a 1536-dimensional FFN.

For SE(3)-Transformer \citep{fuchs2020se3transformers}, we use the official implementation and default settings. We use an 8-head equivariant attention backbone with 6 attention layers, followed by a TFN readout and pooling.

\stitle{Details of \model.}
For our proposed \model, we follow \method{ComENet} \citep{wang2022comenet} to extract 3D conformations and chemical bond types as bond-level attributes. For atomic forces, we directly use the original force values as atom-level attribute. 
We use GCN as the backbone, with its hidden dimension set to 64. We use a dual-layer MLP with bottleneck structure as the conditional network, and set the hidden dimension of the conditional network as 32. 

\section{Visualization}\label{sec.app.visulization}
\begin{figure}[t]
\centering
\includegraphics[width=1\linewidth]{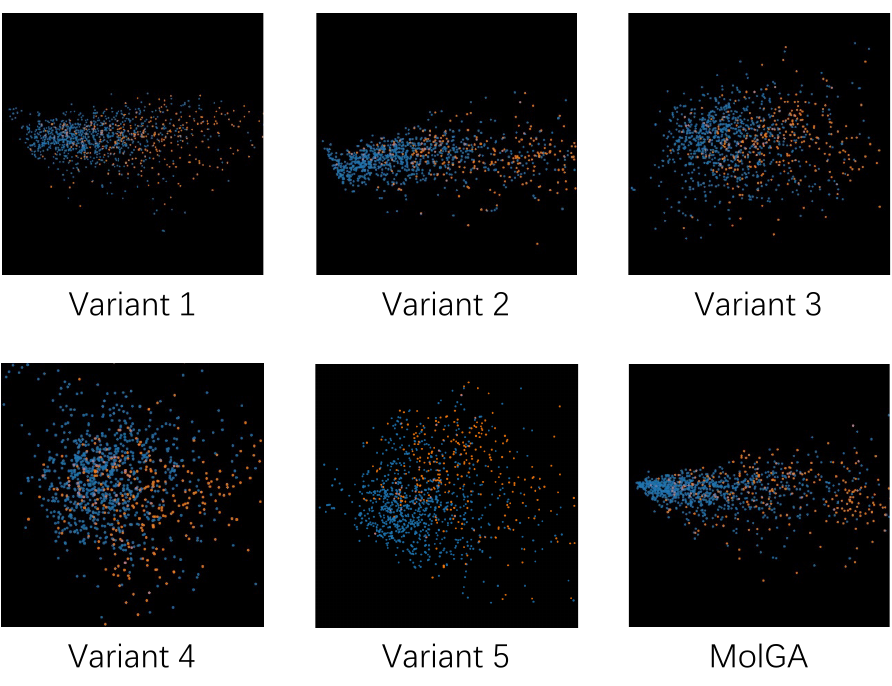}
\vspace{-8mm}
\caption{Visualization of embedding space of atoms.}
\label{fig.visualization}
\end{figure}

To further demonstrate the effectiveness of our core components, we present the atom embedding space on the \textit{BBBP} dataset in Fig.~\ref{fig.visualization}, with different colors indicating distinct atom classes.
We observe that, with the incorporation of molecular alignment and conditional adaptation, atom embeddings from different classes exhibit clear separation. This structure reflects a well-organized latent space shaped jointly by 2D topological information and molecular domain knowledge, highlighting the effectiveness of \model.

\section{GenAI Usage Disclosure}
Large language models (LLMs) were used in the preparation of this manuscript exclusively as language-polishing tools. Specifically, we employed LLMs to refine grammar, improve clarity, and enhance readability of the text. All research ideas, methodologies, experiments, analyses, and conclusions presented in this work are entirely the authors’ own. The LLMs did not contribute to the conception of the research, experimental design, data analysis, or interpretation of results. The authors take full responsibility for the accuracy, originality, and integrity of all scientific content.

\bibliographystyle{ACM-Reference-Format}
\bibliography{references}
\balance
\clearpage
\newpage


\end{document}